\documentclass[letterpaper, 10 pt, conference]{ieeeconf}  

\IEEEoverridecommandlockouts                              

\usepackage {graphicx} 

\usepackage{algorithm}
\usepackage{algorithmic}
\makeatletter
\newcommand{\removelatexerror}{\let\@latex@error\@gobble}
\makeatother
\usepackage{amssymb}
\usepackage{amsmath}
\usepackage{txfonts}

\usepackage{subfigure}
\usepackage{tabularx}  
\usepackage{geometry}  
\usepackage{lipsum} 
\usepackage{array}
\usepackage{multirow}
\usepackage{tabularx}
\usepackage{todonotes}
\usepackage{booktabs}
\usepackage{xcolor}
\title
{\LARGE \bf
Subconscious Robotic Imitation Learning}

\author{Jun Xie$^{*}$, Zhicheng Wang$^{*}$, Jianwei Tan$^{}$, Huanxu Lin$^{}$  and Xiaoguang Ma$^{}$
\thanks{*These authors contributed equally.}
\thanks{Jun Xie and Huanxu Lin are with the Faculty of Robot Science and Engineering, Northeastern University, Shenyang 110819, China. {\{2402187, 2402181\}@stu.neu.edu.cn.}}%
\thanks{Zhicheng Wang, and Jianwei Tan are with the College of Information Science and Engineering, Northeastern University, Shenyang 110819, China. {\{2390108, 2410359\}@stu.neu.edu.cn.}}
\thanks{Xiaoguang Ma is with the College of Information Science and Engineering, the Faculty of Robot Science and Engineering, Northeastern University, Shenyang 110819, China, and the Foshan Graduate School, Northeastern University, Foshan 528311, China. {maxg@mail.neu.edu.cn.}}
\thanks{Corresponding author: Xiaoguang Ma.}}
\begin{document}
\maketitle
\thispagestyle{empty}
\pagestyle{empty}


\begin{abstract}
Although robotic imitation learning (RIL) is promising for embodied intelligent robots, existing RIL approaches rely on computationally intensive multi-model trajectory predictions, resulting in slow execution and limited real-time responsiveness.
Instead, human beings subconscious can constantly process and store vast amounts of information from their experiences, perceptions, and learning, allowing them to fulfill complex actions such as riding a bike, without consciously thinking about each.
Inspired by this phenomenon in action neurology, we introduced subconscious robotic imitation learning (SRIL), wherein cognitive offloading was combined with historical action chunkings to reduce delays caused by model inferences, thereby accelerating task execution. 
This process was further enhanced by subconscious downsampling and pattern augmented learning policy wherein intent-rich information was addressed with quantized sampling techniques to improve manipulation efficiency. Experimental results demonstrated that execution speeds of the SRIL were 100\% to 200\% faster over SOTA policies for comprehensive dual-arm tasks, with consistently higher success rates.

\end{abstract}

\begin{figure*}[t]
    \centering
    \centerline{\includegraphics[width=\linewidth]{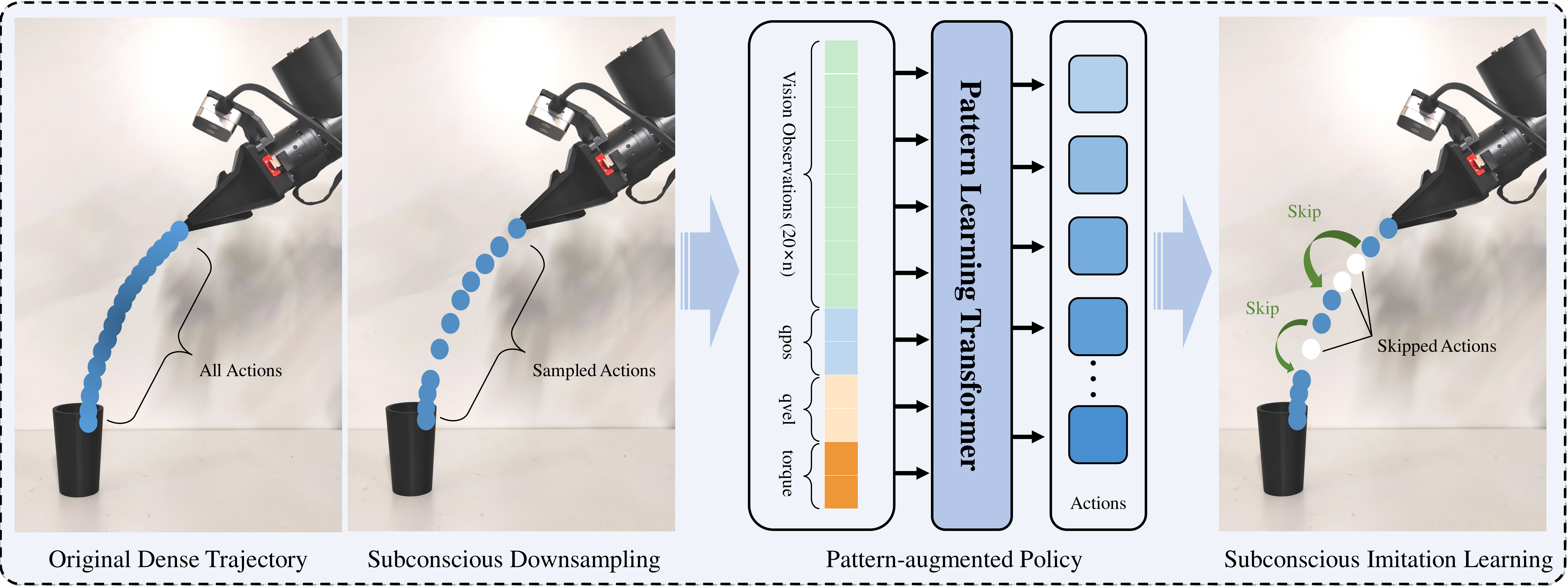}}
    \caption{Overview of the Subconscious Robotic Imitation Learning SRIL framework. \textbf{Left}: Original dense trajectory was subconsciously downsampled to retain key actions. \textbf{Middle}: A transformer-based pattern-augmented learning policy integrated visual observations and subconscious patterns. \textbf{Right}: The policy performed subconscioued imitation rate learning, skipping redundant actions to accelerate task execution. The SRIL highly speeded up execution reduced while preserving performance.}
    
    \label{fig:1}
\end{figure*}


\section{INTRODUCTION}
Robotic Imitation Learning (RIL) has emerged as a powerful data-driven approach for advancing robotic manipulation tasks. \cite{mahmoudi2024leveraging}
By leveraging expert demonstration data, RIL enables robots to autonomously acquire complex control strategies, reducing the reliance on debug-intensive manual programming. \cite{zare2024survey}
This approach has found widespread applications in single-arm, dual-arm robotic manipulation and humanoids scenarios, where RIL allows robots to emulate human operators and perform a wide array of tasks, including grasping \cite{tsarouchi2016human}, assembly \cite{laudante2020human}, and precision manufacturing \cite{caggiano2018digital}.

In RIL, execution speed plays a pivotal role in practical applications, as it directly affects a system's efficiency and effectiveness \cite{hu2024fusion}. 
For example, achieving productivity on par with human performance is essential to enable large-scale replacement of human workers in industrial settings \cite{ravichandar2020recent}. Among the notable advancements in RIL, Action Chunking with Transformers (ACT) and Diffusion Policies (DP) demonstrated significant potential. ACT enhanced task performance by predicting joint positions for future k time steps, effectively shortening the task horizon and reducing cumulative prediction errors \cite{zhao2023learning}. 

In contrast, DP employed conditional denoising diffusion processes to frame visuotactile learning policy as iterative noise optimization \cite{chi2023diffusion}. 
This approach improved the generation of multimodal behaviors and stabilized control in high-dimensional action spaces. 
While these methods bolstered RIL robustness and generalization in complex tasks, their reliance on multi-step predictions and computationally intensive processes greatly increased execution time, thereby undermining real-time responsiveness and operational efficiency \cite{kang2024efficient,zhang2022gddim}. 

Traditional path-planning and control algorithms, such as Rapidly-exploring Random Trees (RRT) \cite{moon2014kinodynamic} and Dynamic Programming (DP) \cite{shin1986dynamic}, improved execution speed in specific scenarios but often suffered from high computational demands and limited scalability to high-dimensional tasks \cite{zhu2020robotic}. 
Furthermore, current methods struggled to adapt to dynamic environmental changes, failing to achieve an optimal balance between precision and speed in such scenarios \cite{karim2024vil}.

Compared to all the above-mentioned RIL approaches, subconscious imitation learning (SIL) of human beings has excellent capabilities on both execution efficiency and accuracy. In fact, the SIL can process and store information from past experience and allow precise task executions without need of predicting each trajectory. The SIL can be facilitated by cognitive offloading, which allows the system to prioritize critical aspects of the task, thereby effectively reducing cognitive load \cite{risko2016cognitive}. Additionally, due to its remarkable ability to recognize patterns in complex environments, the SIL can quickly identify familiar situations and anticipate the necessary actions based on past experience. Furthermore, it continuously adjusts and refines organization based on the outcomes of actions and prior experiences \cite{simpson2024subliminal}.

Inspired by SIL, we proposed a subconscious robotic imitation learning (SRIL) to framework to address two primary factors contributing to their low execution efficiency, \textit{i.e.,} data redundancy and redundancy in inference-execution. 
As illustrated in Fig. \ref{fig:1}, the key contributions of our approach were summarized as follows:

\begin{itemize}
\item \textbf{Subconscious Downsampling}: We proposed a subconscious downsampling method that operated between fully conscious keypoint extraction and dense trajectory recording. Utilizing motion intention data (e.g., joint velocities, gripper torques), it efficiently removed redundant trajectory information, enhancing both efficiency and generalizability.

\item \textbf{Pattern-augmented Learning Policy}: We proposed a concise neural network model incorporating pattern-augmented information, enhancing system robustness and success rates with the help of subconscious downsampling.

\item \textbf{Subconscious Imitation Learning}: Inspired by human beings subconscious, we proposed a subconscious imitation learning mechanism that leveraged historical chunked trajectories to offload routine sub-trajectories. This approach reduced policy computation overhead in well-learned segments, thereby accelerating task execution. 
\end{itemize}

Comprehensive simulation and real robot experiments illustrated that execution speeds of the SRIL were 100\% to 200\% faster over SOTA policies for comprehensive dual-arm tasks, with consistently higher success rates, showing its great potential as a backbone to robotic imitation learning.

\section{RELATED WORK}
\begin{figure*}[t]
    \centering
    \centerline{\includegraphics[width=\linewidth]{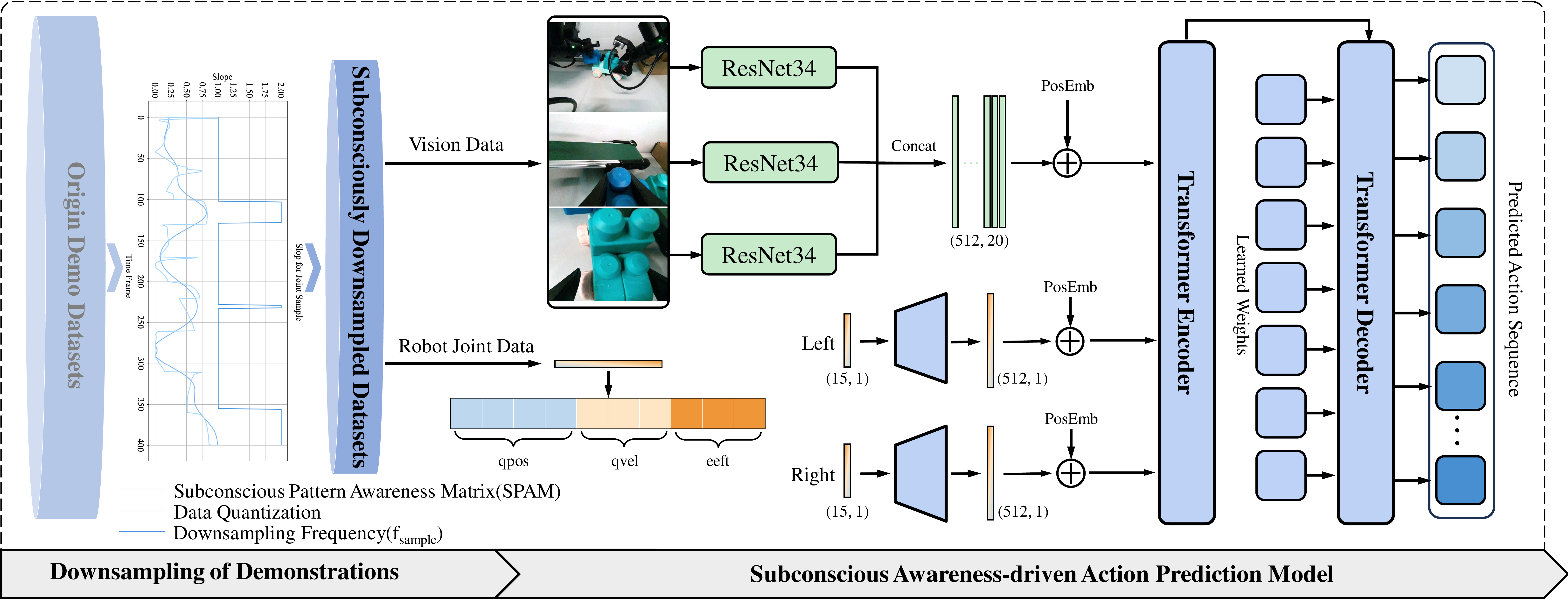}}
    \caption{Subconscious Downsampling and Subconscious Pattern-Augmented Learning Policy. \textbf{ Left}: The demonstration datasets were downsampled via subconscious pattern recognition to create subconscious downsampled datasets, which train the Pattern-Driven Action Prediction Model. \textbf{ Right}: The model combines visual and joint data through ResNet encoders and a Transformer architecture for precise manipulation prediction.}
    \label{fig:2}
\end{figure*}

\subsection{Advances and Challenges in Robotic Imitation Learning (RIL)}
RIL made significant strides in improving task accuracy and control stability for robotic applications\cite{zhang2021reinforcement}. Approaches such as RVT \cite{goyal2023rvt} and PerAct \cite{shridhar2023perceiver} leveraged voxel-based 3D modeling to enhance spatial accuracy, enabling precise identification of target regions in complex environments.

Similarly, systems like VoxPoser \cite{huang2023voxposer} and ReKep \cite{huang2024rekep} integrated advanced multimodal techniques, including tools such as ChatGPT-4V, to optimize trajectory planning, resulting in improved motion accuracy and spatial detection.
The Action Chunk Transformer (ACT) further advanced RIL by introducing multi-step trajectory prediction, which reduced error accumulation and enhanced task execution stability. \cite{george2023one}.

Meanwhile, Diffusion Policies (DP) emploied denoising-based techniques to improve precision \cite{wang2024one,liu2024enabling}.
Methods such as UMI \cite{chi2024universal} integrated VSLAM \cite{fuentes2015visual} during data pre-processing to bolster training robustness. 
Although these advancements significantly improved RIL, their applications remained highly constrained by the limited performance capabilities in real-world environments.

In fact, SOTA RIL approaches like ACT and DP often involved computationally intensive mechanisms, which led to delays in decision-making and execution, particularly in high-frequency dynamic tasks where responsiveness was critical \cite{yang2024vq}. Addressing this challenge required innovative RIL frameworks allowing fast execution in real-time scenes with high accuracy.
\subsection{Execution Efficiency of Robotic Manipulation}
Researchers proposed various strategies to enhance execution efficiency of robotic systems, including path planning and control algorithms. Techniques such as Rapidly-exploring Random Trees (RRT) \cite{zhu2020robotic} and Dynamic Programming (DP) \cite{karim2024vil} improved response times in low-dimensional tasks but were hindered by high computational costs in complex and high-dimensional environments, limiting their effectiveness in dynamic scenarios \cite{xu2022light}. 
Adaptive sampling methods were introduced to reduce redundant data and enhance computational efficiency \cite{hua2021learning}. However, these methods often lacked the flexibility required for real-time tasks with high-frequency, making them less effective in dynamically changing environments.

Recently, the integration of deep reinforcement learning with multi-objective trajectory optimization, as demonstrated by Zhang et al. \cite{zhang2023multi} in sparse reward environments for robotic manipulation, showed promise for improving efficiency in complex scenarios. 
Nevertheless, the highly computational demands of these methods still posed challenges to their feasibility and scalability in real-time applications \cite{zhu2023intelligent}. Similarly, physical modeling-based path optimization, commonly applied to pipeline tasks, achieved efficiency gains for fixed paths but struggled to adapt to unstructured and dynamic environments. 
In fact, the trade-off between computational efficiency and execution speed remained an unresolved challenge in real-time robotic scenes \cite{khanna2023path}.
\subsection{Biological-based Robotic Manipulations}
In dynamic environments, biological organisms utilized memory and feedback mechanisms to enhance decision-making speed and adaptability \cite{iida2016biologically}. Drawing inspiration from these principles, researchers proposed innovative RIL algorithms. For example, Guo et al \cite{guo2024remember}. introduced the Fragmented Memory Trajectory Prediction (FMTP) model, which leveraged discrete storage to reduce data redundancy, thereby increasing trajectory generation efficiency in complex tasks. Similarly, Debat et al \cite{debat2021event}. developed a Synaptic Neural Network (SNN) that improved trajectory prediction and motion decision-making, enhancing robots' responsiveness to dynamic conditions. While these approaches made progresses in trajectory generation and execution efficiency, achieving real-time adaptability and superior performance in high-dimensional remained an ongoing challenge \cite{iida2016biologically,parisotto2017neural}.
\section{METHODS}

\subsection{Subconscious Pattern Awareness and Downsampling}
To train efficient imitation learning models, we collected a dataset $\mathcal{D} = \{\tau_0, \tau_1, \dots, \tau_N\}$ comprising $N$ expert-demonstrated trajectories. Each trajectory was formalized as
\begin{equation}\label{eq:001}
    \tau_n = \{(O_{(n,t)}, S_{(n,t)}, A_{(n,t)})\}, \quad n \in [0, N],\ \ t \in [0, T],
\end{equation}
where $O_{(n,t)}$ represented visual observations, $S_{(n,t)}$ denoted sensory data, and $A_{(n,t)}$ encapsulated action data. Here, $n$ indexed the trajectory, and $t$ denoted the time steps within a trajectory. The sensory data $S_{(n,t)}$, which included joint positions, joint velocities, and end-effector torques, were represented as
\begin{equation}\label{eq:002}
    S_{(n,t)} = \{qpos_{(n,t)}, qvel_{(n,t)}, eeft_{(n,t)}\},
\end{equation}
where $qpos_{(n,t)}$, $qvel_{(n,t)}$, and $eeft_{(n,t)}$ corresponded to joint positions, joint velocities, and end-effector torques, respectively. The action data $A_{(n,t)}$, which guided the robot's movements, were represented as joint positions and velocities
\begin{equation}\label{eq:003}
    A_{(n,t)} = \{\alpha_{pos_{(n,t)}}, \alpha_{qvel_{(n,t)}}\}.
\end{equation}

Where, $\alpha_{pos_{(n,t)}}$ denoted the target joint positions, and $\alpha_{qvel_{(n,t)}}$ represented the target joint velocities. These parameters provided precise control instructions and determined robots' abilities to accomplish manipulation tasks.
To retain critical patterns, which could be efficiently recognized by subconscious, and to reduce redundancy during data preprocessing, we designed an subconscious pattern awareness matric (SPAM) $I_t$ to quantify the impact of joint velocities and torques on task-relevant patterns \textit{i.e.,}
\begin{equation}\label{eq:004}
    I_t = \begin{bmatrix} 
    \mathcal{N}\left(\frac{1}{N} \sum_{n=1}^N \alpha_{qvel_{(n,t)}} \right) ; 
    \mathcal{N}\left(\frac{1}{N} \sum_{n=1}^N eeft_{(n,t)} \right)
    \end{bmatrix} \cdot \omega,
\end{equation}
where $\mathcal{N}(\cdot)$ was a normalization function used to eliminate dimensional disparities among different joint velocities and torques, and $\omega$ was a weight vector that reflected the relative importance of each joint. This metric prioritized preserving task-critical patterns during downsampling, effectively reducing data redundancy while enhancing prediction abilities of the model. Based on the SPAM $I_t$, the downsampling frequency $f_{\text{simple}}(t)$ was defined as
\begin{equation}\label{eq:005}
f_{\text{simple}}(t) = \frac{f_d}{\lfloor M \cdot \mathcal{L}_{fc}(I_t) \rfloor + 1},
\end{equation}
where $f_d$ was the original sampling frequency, $f_m$ was the target minimum sampling frequency, $M = \frac{f_d}{f_m}$ was a temporal scaling factor, $\mathcal{L}{fc}(\cdot)$ was a Butterworth filter used to smooth $I_t$ while preserving its low-frequency characteristics, and $\lfloor \cdot \rfloor$ denoted the floor operation, ensuring integer sampling indices.
Using the above dynamic downsampling frequency, the original trajectory $\tau_n$ was downsampled into a subconscious trajectory $\tau_{\text{ds},n}$, \textit{i.e.,} subconscious trajectory with sampling indices determined by
\begin{equation}\label{eq:006}
\mathcal{I}_{\text{ds}} = \{k \mid k = \sum_{j=0}^{n-1} \frac{f_d}{f_{\text{ds}}(j)}, \, k < T, \, n \in \mathbb{Z}^+\}.
\end{equation}
The subconscious trajectory $\tau_{\text{ds},n}$ was represented as
\begin{equation}\label{eq:007}
\tau_{\text{ds},n} = \{(O_{(n,k)}, S_{(n,k)}, A_{(n,k)}) \mid k \in \mathcal{I}_{\text{ds}}\}.
\end{equation}

Finally, the downsampled dataset were given by
\begin{equation}\label{eq:008}
\mathcal{D}_{\text{ds}} = \{\tau_{\text{ds} (0)}, \tau_{\text{ds}(1)}, \dots, \tau_{\text{ds} (N)}\}.
\end{equation}

As illustrated in Fig. \ref{fig:2}, during the ``subconscious Downsampling" phase, the SPAM $I_t$ was employed to quantify the influence of joint velocities and end-effector torques on task-relevant patterns. By leveraging Eq.~\eqref{eq:004}, $I_t$ was calculated and combined with a Butterworth filter to retain essential low-frequency features while dynamically adjusting the downsampling frequency $f_{\text{ds}}$ as defined in Eq.~\eqref{eq:005}. The downsampled data were reorganized into a subconscious downsampled dataset, significantly reducing redundant information while preserving task-critical patterns. This process established a solid foundation for efficient training of subsequent models.

\subsection{Subconscious Pattern-Augmented Learning Policy}
Based on the fact that subconscious could excellently recognize trajectory patterns in manipulation and predict future trajectories via past experience, this section introduced an action prediction model based on the Transformer architecture to enhance generalization and reasoning capabilities in complex robotic tasks by modeling the key patterns preserved in downsampled trajectories. By integrating visual observations $O_{(n,k)}$ and state data $S_{(n,k)}$, the model could predict target actions $A_{(n,k)}$, facilitating efficient imitation of robotic manipulation tasks.

For a dual-arm robotic manipulation task, each subconscious downsampled trajectory $\tau_{(\text{ds},n)}$ was processed as follows: the visual observations $O_{(n,k)} \in \mathbb{R}^{640 \times 480}$ were passed through a pretrained convolutional neural network (ResNet34) to extract feature vectors $F_{I} \in \mathbb{R}^{20 \times 512}$. These visual features $F_{I}$ were concatenated with state data $S_{(n,k)}$ and then projected into a unified feature space through a multilayer perceptron (MLP)
\begin{equation}\label{eq:009}
X_{(n,k)} = \text{MLP}\left(\text{concat}(F_{I}, S_{(n,k)})\right), \quad X_{(n,k)} \in \mathbb{R}^{512},
\end{equation}
where $S_{(n,k)} = \{qpos_{(n,k)}, qvel_{(n,k)}, eeft_{(n,k)}\}$ represented the robot's joint positions $qpos_{(n,k)} \in \mathbb{R}^{14}$, joint velocities $qvel_{(n,k)} \in \mathbb{R}^{14}$, and end-effector forces/torques $eeft_{(n,k)} \in \mathbb{R}^{2}$. 
The input feature sequence $\{X_{(n,k)}\}_{k=1}^{T_{\text{ds}}}$, after positional embedding, was processed by a Transformer encoder to capture global contextual information
\begin{equation}\label{eq:010}
H_{(n)} = \text{Transformer}_{\text{Encoder}}(X_{(n,k)} + \text{PosEmb}(k)),
\end{equation}
where $H_{(n)} \in \mathbb{R}^{T_{\text{ds}} \times 512}$ represented the latent representation of the trajectory, and $T_{\text{ds}}$ was the trajectory length.
In the action generation phase, the Transformer decoder predicted the target action for the current time step based on the encoder output $H_{(n)}$ and the action from the previous time step $A_{(n,k-1)}$
\begin{equation}\label{eq:011}
A_{(n,k)} = \text{Transformer}_{\text{Decoder}}(H_{(n)}, A_{(n,k-1)}),
\end{equation}
where $A_{(n,k)} = \{\alpha_{pos_{(n,k)}}, \alpha_{qvel_{(n,k)}}\}$ represented the predicted target joint positions and velocities. 
The model was trained to minimize the mean squared error between the predicted actions and the ground truth actions, using the following loss function
\begin{equation}\label{eq:012}
\mathcal{L}_{\text{action}} = \frac{1}{N \cdot T_{\text{ds}}} \sum_{n=1}^N \sum_{k=1}^{T_{\text{ds}}} \|A_{(n,k)} - A_{(n,k)}^{\text{true}}\|_2^2,
\end{equation}
where $N$ denoted the number of trajectories.

As shown in Fig. \ref{fig:2}, the proposed subconscious pattern-augmented learning policy integrated visual observations $O_{(n,k)}$ and state data $S_{(n,k)}$ to generate target action sequences $A_{(n,k)}$. Visual data, such as RGB images from the robot's operational environment, were processed through pretrained ResNet34 networks to extract features $F_I$, resulting in a $512 \times 20$ feature vector.
These visual features were concatenated with robot state data $S_{(n,k)}$, which included joint positions ($qpos$), joint velocities ($qvel$), and end-effector forces/torques ($eeft$). The concatenated features were then projected into a unified feature space using the MLP, as described in Eq.~\eqref{eq:009}.

Subsequently, the input feature sequence ${X_{(n,k)}}$ was processed by a Transformer encoder to capture global contextual information across the trajectory (Eq.~\eqref{eq:010}). During the decoding phase, a Transformer decoder combined the encoded output $H_{(n)}$ with the predicted action from the previous time step $A_{(n,k-1)}$ to generate the current target action $A_{(n,k)}$ (Eq.~\eqref{eq:011}). This Transformer-based architecture effectively modeled the key patterns retained in the downsampled trajectories, enabling high-precision action prediction for complex manipulation tasks.

\subsection{Action Execution Strategy Based on Subconscious Robotic Imitation Learning (SRIL)} 
During the action execution phase of a robotic manipulator, the system predicted future action sequences $a_t = \{a_{t}[t+1], a_{t}[t+2], \dots, a_{t}[t+K]\}$ based on current observation $o_t$, where $K$ denoted the length of the action block. The executed action at time $T+1$ was computed as a weighted cumulative prediction of multiple historical predictions using an exponential weighting scheme
\begin{equation}\label{eq:013}
A_{T+1} = \frac{\sum_{k=0}^{K-1} a_{T-k}[T+1] \cdot \exp(-m \cdot k)}{\sum_{k=0}^{K-1} \exp(-m \cdot k)},
\end{equation}
where $\exp(-m \cdot k)$ was a temporal weighting factor that balanced the importance of recent versus earlier predictions.

To realize SRIL, we introduced the concept of cognitive offloading in cognitive science. Cognitive offloading reduced cognitive load by relying on external mechanisms or prior experiences, improving task efficiency without continuous real-time decision-making. Inspired by biological organisms, which offloaded routine tasks to focus on critical decisions, robotic systems could assess readiness for offloading using the Cognitive Offloading Readiness (COR) metric.
To evaluate the readiness for cognitive offloading, the Cognitive Offloading Readiness (COR) metric was calculated based on the overlapping joint position information when the historical prediction trajectories exhibited high consistency, \textit{i.e.,} 
\begin{equation}\label{eq:014}
COR_{T+1} = \sum_{j=1}^{n} \texttt{std}\left( \frac{a_{(:,j)}[T+1] - \mu_j}{\sigma_j} \right),
\end{equation}
where $a_{(:,j)}[T+1]$ represented the predicted value sequence for joint $j$, and $\mu_j$ and $\sigma_j$ were the mean and standard deviation of the sequence, respectively. $n$ denoted the total number of joints. The $COR_{T+1}$ quantified prediction consistency and determined whether cognitive offloading could be initiated. The cognitive offloading threshold ($COT$) determined when offloading was triggered, and the minimum cognitive engagement duration ($mced$) ensured a sufficient history of decisions before offloading occurred. The system skipped subsequent model inference (denoted as \textit{skip}) if the following conditions were met
\begin{equation}\label{eq:015}
A_{T+1} = 
\begin{cases} 
    \text{skip}, & COR_{T+1} > COT \land \phi, \\ 
    \text{infer}, & \text{otherwise},
\end{cases} 
\end{equation}
where $\text{skip}$ signified bypassing subsequent frame computations, $\text{infer}$ indicated the continuation of conscious inference, $\phi= \text{len}(a_{(:)}[T+1]) \geq mced$, and $\text{len}(a_{(:)}[T+1])$ was the length of the available historical trajectory.

The duration of offloading was defined as $\min(MCOD, n)$, where $n$ represented the length of the historical prediction trajectory. In simulation environments, the maximum cognitive offloading duration ($MCOD$) was typically set to $K$. To ensure the reliability of the $COR$ metric, the minimum cognitive engagement requirement ($mced$) had to be satisfied, guaranteeing the availability of adequate historical decision data before initiating cognitive offloading.
\begin{figure}
	\centering
	\includegraphics[width=1.0\linewidth]{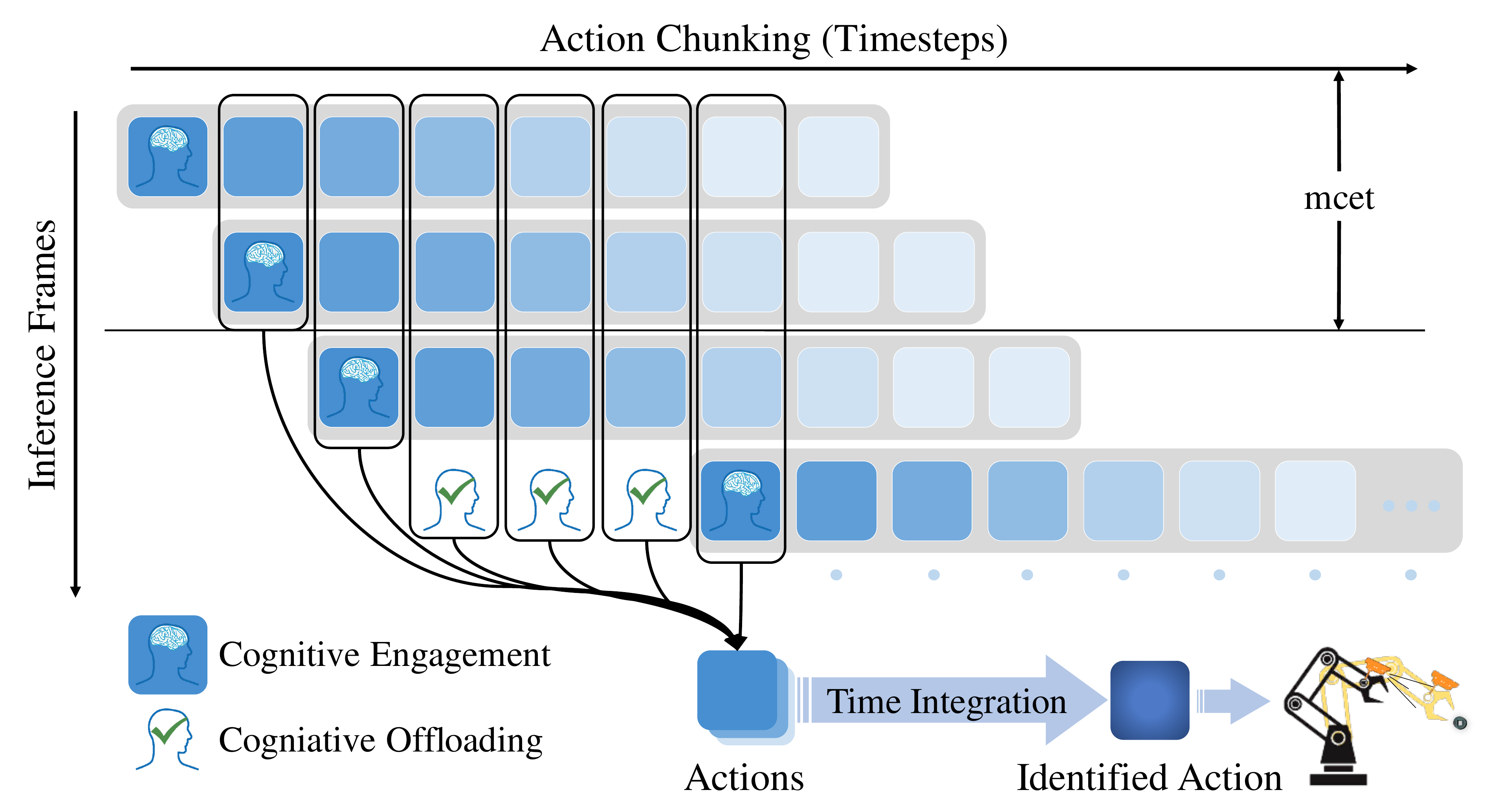}
	\caption{Cognitive Offloading for Subconscious Imitation Learning: Each inference frame generates an action chunking sequence, triggering cognitive offloading when the COR exceeds the COT, enabling efficient subconscious imitation through action time integration.}
	\label{fig:3}
\end{figure}
\begin{figure*}[!htb]
	\centering
	\centerline{\includegraphics[width=\linewidth]{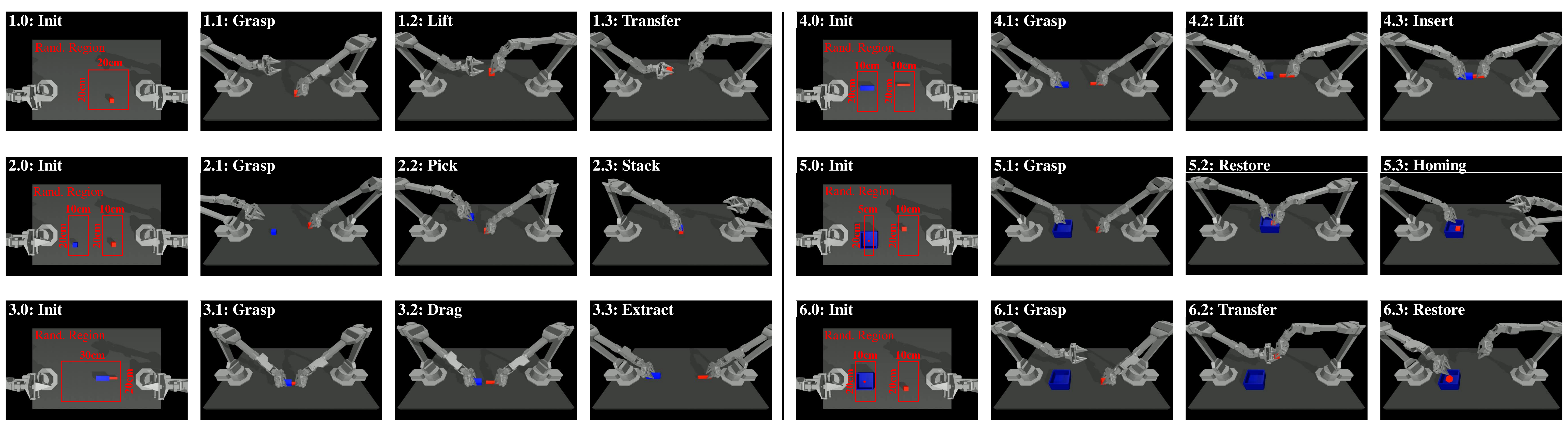}}
	\caption{Overview of the simulation tasks. 
		\textbf{Task 1 Cube Transfer}: The right arm grasped (1.1) and lifted the red cube (1.2), then transferred it to the left arm (1.3). 
		\textbf{Task 2 Bimanual Stack}: The right arm grasped and lifted the red cube (2.1), and the left arm grasped the blue cube (2.2). Then the right arm put the red cube in the middle of the table and the left arm put the blue cube on top of the red cube (2.3).
		\textbf{Task 3 Bimanual Extraction}: The left arm grasped the blue socket and the right arm grasped the red peg (3.1). Both dragged and extracted the inserted parts (3.2) and placed them on the table (3.3).
		\textbf{Task 4 Bimanual Insertion}: The left arm grasped the blue socket and the right arm grasped the red peg (4.1). Both lifted (4.2) and inserted the red peg into the blue socket (4.3). 
		\textbf{Task 5 Bimanual Restore}: The left arm picked up the blue box and the right arm picked up the red cube (5.1). Then they lifted and placed the red cube into the blue box above the table (5.2). After that, the left arm put the box back (5.3).
		\textbf{Task 6 Transfer and Restore}: The right arm grasped (6.1) and transferred the red cube to the left arm (6.2). Then the left arm placed the red cube into the blue box (6.3).
	}
	\label{fig:4}
\end{figure*}
\begin{table*}[!htb]
	\centering
	\caption{Simulation Results.}\label{tab1}
	\begin{tabular}{@{}p{2.5cm}cp{0.8cm}cp{0.8cm}cp{1.5cm}cp{0.8cm}cp{0.8cm}cp{1.5cm}cp{0.8cm}cp{0.8cm}cp{1.5cm}c@{}}
		\toprule
		\multirow{3}{*}{Policy} & \multicolumn{3}{c}{1:Cube Transfer} & \multicolumn{3}{c}{2:Bimanual Extraction} & \multicolumn{3}{c}{3:Bimanual Stack} \\
		\cmidrule(lr){2-4} \cmidrule(lr){5-7} \cmidrule(lr){8-10}
		& Lift & Transfer & Cost Time & Grasp & Extract & Cost Time & Pick & Stack & Cost Time\\ 
		\midrule
		Diffusion Policy & 25.25\% & 10.20\% & 20.15s & 41.66\% & 45.45\% & 24.76s & 23.81\% & 13.69\% & 39.24s\\
		ACT (No Integration) & 94.34\% & 92.59\% & 9.99s & 98.04\% & 89.29\% & 10.88s & \textbf{94.34\%} & 90.91\% & 24.55s \\
		ACT & 94.34\% & 94.33\% & 12.46s & 98.04\% & 94.33\% & 14.85s & 92.59\% & 92.59\% & 33.20s \\
		SRIL (W/O SIL) & \textbf{96.15\%} & \textbf{96.15\%} & 8.77s & \textbf{98.04\%} & \textbf{98.04\% }& 10.79s & 92.59\% & 92.59\% & 24.84s \\
		SRIL & 94.34\% & 94.34\% & \textbf{4.51s} & \textbf{98.04\%} & \textbf{98.04\%} & \textbf{4.76s} & \textbf{94.34\%} & \textbf{94.34\%} & \textbf{12.26s}  \\
		
	\end{tabular}

	\begin{tabular}{@{}p{2.5cm}cp{0.8cm}cp{0.8cm}cp{1.5cm}cp{0.8cm}cp{0.8cm}cp{1.5cm}cp{0.8cm}cp{0.8cm}cp{1.5cm}c@{}}
		\toprule
		\multirow{3}{*}{Policy} & \multicolumn{3}{c}{4:Bimanual Insertion} & \multicolumn{3}{c}{5:Transfer and Restore} & \multicolumn{3}{c}{6:Bimanual Restore} \\
		\cmidrule(lr){2-4} \cmidrule(lr){5-7} \cmidrule(lr){8-10}
		& Lift & Insert & Cost Time & Transfer & Restore & Cost Time & Restore & Homing & Cost Time \\ 
		\midrule
		Diffusion Policy & 15.63\% & 2.41\% & 23.78s & 23.25\% & 3.98\% & 29.76s & 4.76\% & 0.00\% & 26.42s  \\
		ACT (No Integration) & 86.36\% & 45.45\% & 15.82s & 42.42\% & 37.88\% & 24.29s & 48.24\% & 43.85\% & 20.47s \\
		ACT & 89.77\% & 28.41\% & 20.57s & 53.76\% & 53.76\% & 31.70s & \textbf{59.57\%} & 53.19\% & 23.96s \\
		SRIL (W/O SIL) & 92.13\% & 56.18\% & 13.44s & 54.64\% & \textbf{54.64\%} & 26.18s & 57.14\% & \textbf{55.56\%} & 23.43s \\
		SRIL & \textbf{92.21\%} & \textbf{64.94\%} & \textbf{9.33s} & \textbf{55.21\%} & 52.08\% & \textbf{14.47s}  & 57.55\% & 53.76\% & \textbf{12.11s} \\
		\bottomrule
	\end{tabular}
	
	\label{tab:performance_sim}
\end{table*}
\section{EXPERIMENTS}
Our experiments systematically evaluated the subconscious imitation learning (SIL) framework across three key aspects \textit{i.e.,} simulation-based efficacy, parameter sensitivity, and real-world applicability.
In six simulation tasks, we demonstrated that the framework significantly improved task learning outcomes by reducing execution time while maintaining high accuracy.
Parameter sensitivity analysis revealed how key factors influenced performance and scalability, offering insights into the mechanisms enabling SIL.
The framework was further validated in three industrial assembly and disassembly tasks, where it exhibited robust performance under complex conditions.
\subsection{Simulation Experiments}

\subsubsection{Environments Setup}
The experiments were conducted on six dual-arm tasks within the MuJoCo simulation environment, \textit{i.e.,} Cube Transfer, Bimanual Stack, Bimanual Extraction, Bimanual Insertion, Transfer and Restore, and Bimanual Restore.
These tasks featured various levels of randomized conditions and critical point rewards, as depicted in Fig. \ref{fig:4}.
The simulation environment included a table setup with two ViperX 300 robotic arms, each equipped with six degrees of freedom (DoF) joints and parallel pincer clamps.
To ensure comprehensive scene coverage, a top-down camera was utilized for capturing visual input.

\begin{figure*}[t]
    \centering
    \includegraphics[width=1.0\linewidth]{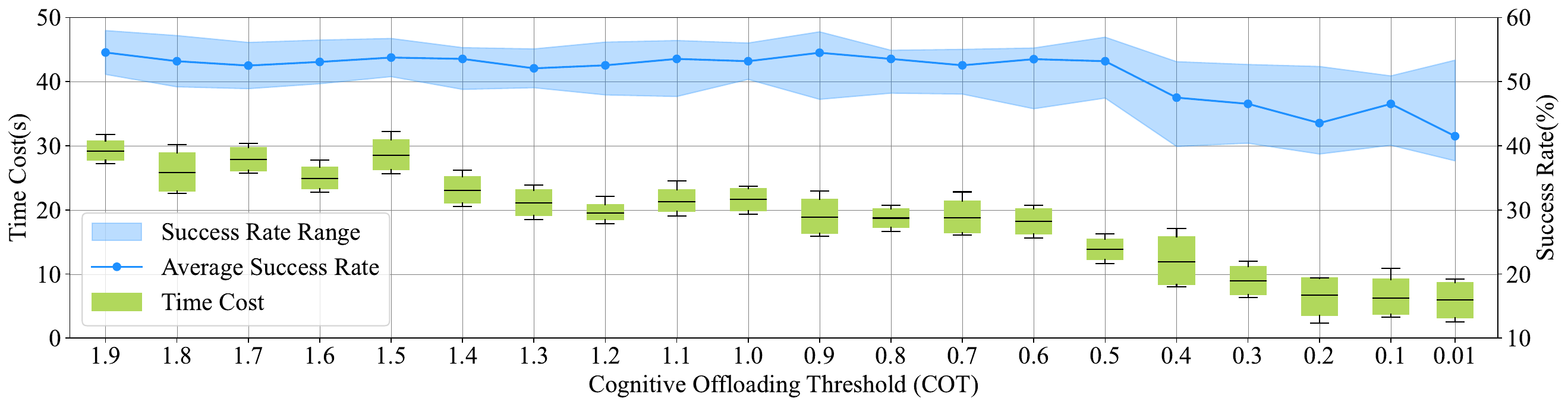}
    \caption{Effects of the COT on Subconscious Robotic Imitation Learning (SRIL).}
    \label{fig:5}
\end{figure*}

To ensure consistency in dataset and inference parameters, the simulation experiments were performed within a controlled physical environment.
The computational setup utilized for training, inference, and rendering comprised a laptop equipped with an Intel i9-13900 CPU and a RTX 3080Ti GPU. 
Task execution time was accurately measured by maintaining a 20-minute interval between inference sessions, during which CPU and GPU temperatures were carefully regulated to stay below $50~^\circ\mathrm{C}$. 
The ambient room temperature was maintained at approximately $10~^\circ\mathrm{C}$ with ventilation open and power supply stabilized.

\subsubsection{Trainning Setup}
To evaluate training efficiency, the Transformer-based autoregressive strategy was trained for 2,000 steps.
This was sufficient for convergence, with the action block size fixed at 100.
The diffusion-based strategy underwent training for 200,000 steps, employing 100 iterations during the diffusion process and 8 time steps per inference.
The optimal strategy was selected for subsequent evaluation.
Each task was tested for $N$ times using a fixed dataset, with a minimum of 50 successful trials.
For the successful trials, the average time from the initial inference to task completion was recorded.
To enhance statistical reliability, the results were averaged over three independent tests conducted with different random seeds.

\begin{figure*}[t]
    \centering
    \centerline{\includegraphics[width=\linewidth]{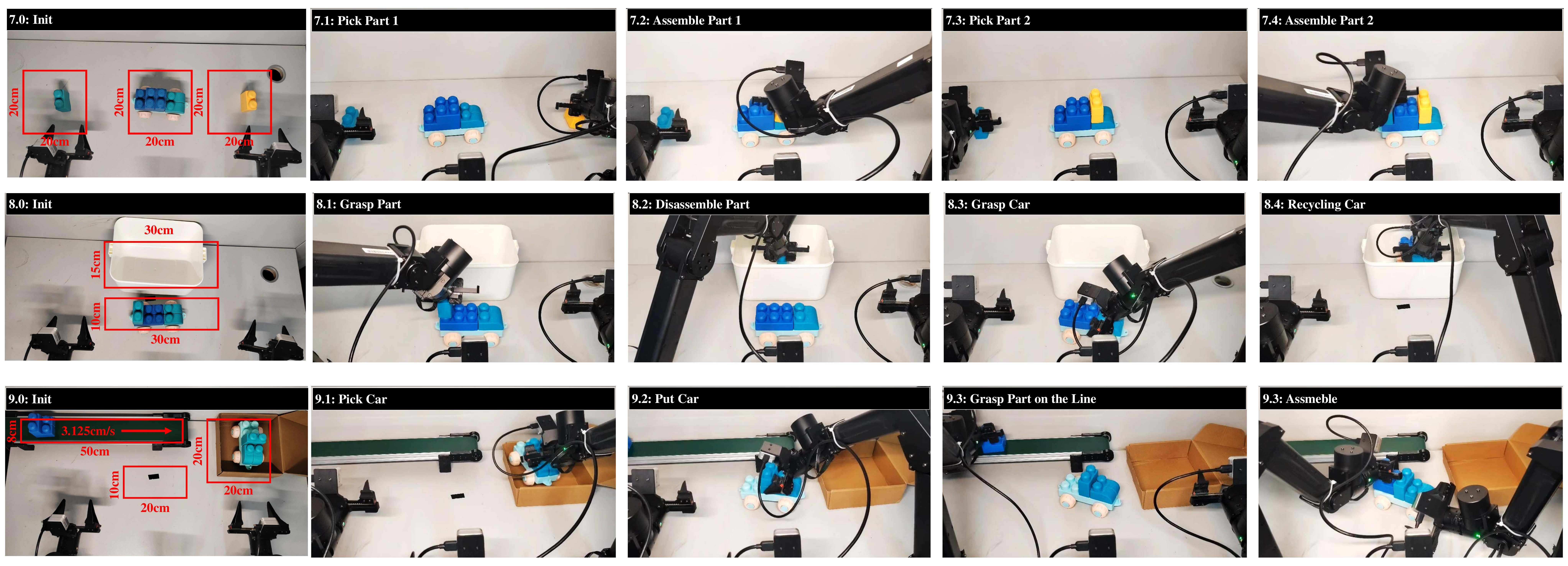}}
    \caption{Overview of real robot tasks. 
    \textbf{Task 7 Assembly}: The right arm grasped (7.1) and mounted the yellow block to the upper counterpart of the block cart (7.2), and then the left arm grasped the dark green block (7.3) and mounted it to the vacant position on the block cart (7.4).
    \textbf{Task 8 Disassembly}: The left arm grasped the extra dark green blocks on the cart (8.1) and placed them in the box (8.2). The right arm grasped the base of the block cart (8.3) and lowered the whole thing into the block box (8.4).
    \textbf{Task 9 Assembly line}: The right arm grasped the trolley to be assembled from the carton (9.1) and placed it stably on the work board (9.2). The left arm waited for the trolley parts on the assembly line and grasped them (9.3), and finally installed them on the trolley's defective parts until the trolley production was completed (9.4).}
    \label{fig:7}
\end{figure*}

\begin{table*}[!htb]
    \centering
    \caption{Real Robot Results.}\label{tab2}
    \begin{tabular}{@{}p{2.5cm}cp{0.8cm}cp{0.8cm}cp{1.5cm}cp{0.8cm}cp{0.8cm}cp{1.5cm}cp{0.8cm}cp{0.8cm}cp{1.5cm}c@{}c}
        \toprule
        \multirow{3}{*}{Policy} & \multicolumn{3}{c}{7:Assembly} & \multicolumn{3}{c}{8:Disassembly} & \multicolumn{3}{c}{9:Assembly line} \\
        \cmidrule(lr){2-4} \cmidrule(lr){5-7} \cmidrule(lr){8-10}
        & Part1 & Part2 & Cost Time & Part & Car & Cost Time & Car & Assmeble & Cost Time\\ 
        \midrule
        ACT & 68.96\% & 47.61\% & 49s & 74.07\% & 51.28\% & 48s & 66.67\% & \textbf{0\%} & $\infty$ \\
        SRIL  & 64.51\% & 52.63\% & \textbf{22s} & 74.07\% & 57.14\% & \textbf{20s} & 60.60\% &\textbf{41.66\%} & \textbf{20s}  \\
        \bottomrule
    \end{tabular}
\end{table*}

\subsubsection{Results and Analysis}
As shown in Table \ref{tab1}, our subconscious robotic imitation learning (SRIL) method achieved the fastest execution speed while maintaining comparable success rates, demonstrating its superior efficiency.
SRIL (W/O SIL) utilized Subconscious Pattern-Aware Downsampling and a subconscious Pattern-Driven Action Prediction Model, but without Subconscious Imitation learning.
This approach reduced the inference time by an average of 31.4\% compared to the ACT strategy with temporal integration.
In contrast, diffusion-based probabilistic strategies, which involved repetitive diffusion and conditional denosing processes during both training and inference, exhibited significantly lower efficiency, with an average execution time 1.6 times longer than that of ACT (No Integration)in 16 action sequences of prediction and direct execution. The sequences generated by this policy are fully included in the training process, making it inefficient to determine an optimal prediction sequence length for specific tasks. Furthermore, incorporating time integration into the diffusion policy, combined with its inherently complex inference time, risks leading to ineffective or "zombie" actions.

SRIL incorporated subconscious execution based on pattern-augmented policy and achieved the best overall performance across six comprehensive two-arm simulation tasks.
While maintaining success rates comparable to SOTA policy such as ACT, it significantly reduced execution time, with an average improvement of 147\% across the six tasks.
For instance, in the two-arm extraction task, the average completion time was reduced to 4.76 seconds, representing a 212\% improvement.
Notably, this execution speed up did not compromise accuracy. In fact, success rates of 94.34\% for cube transfer and 98.04\% for double-arm extraction were obtained, demonstrating superior performance in complex tasks.

During task execution, SRIL dynamically accelerated within a subconscious mimicry state by offloading conscious decision-making and bypassing irrelevant long motion trajectories.
Simultaneously, it autonomously decelerated near target objects, enabling the model to make additional decisions for precise actions.
This adaptive balance between speed and robustness demonstrated its effectiveness in complex scenarios.

Specific analysis focusing on Cognitive Offloading Threshold (COT) was done on the challenging Transfer and Restore task to explore differences between SRIL and ACT models with or without temporal integration, as shown in Fig. \ref{fig:5}. 
The experiments utilized an Maximum Cognitive Offloading Duration (MCOD) of 100, equivalent to the size of Action Chunking, along with the recommended MCOD of 10.

The results indicated that as the COT decreased, the robotic agent’s task execution speed progressively increased.
However, once the COT fell below a critical threshold, the agent became excessively relaxed during task execution.
This over-relaxation accelerated task completion but compromised stability, increasing the failure rate of entirely subconscious task execution.
While this subconscious execution stabilized certain tasks, it led to unpredictable behaviors in others, elucidating the need for careful calibration of the COT to balance speed and stability.
Executing tasks on physical robots under these conditions posed greater risks.
To mitigate these risks, it was recommended to set the MCOD to the default length of Action Chunking as a protective parameter.
For practical deployment, an initial MCOD value of 5 was suggested.
It could then be gradually increased based on the specific specifications and accuracy requirements of the robot.
This approach ensured safer execution while maintaining flexibility for fine-tuning in line with the capabilities of the real robotic system.

\subsection{Real Robot Experiments}

\subsubsection{Real Setup}
The experimental setup consisted of two ARX L5 robotic arms and a VR data acquisition system with high stability, accuracy, and full SDK support.
The system was equipped with three Realsense D405 depth cameras \textit{i.e.,} two were mounted on manipulator wrists to capture hand views, while the third was fixed on the tabletop to provide a third-person perspective of the robot's operation.
To facilitate ergonomic data acquisition during VR teleoperation, the robotic arms, along with their end-mounted cameras, were positioned at the side of the experimental table, with the workspace defined as the desktop area in front.

The hardware used for data acquisition, training, and inference remained consistent with that employed in the simulation experiments.
Since the benchmark ``Diffusion Strategy" demonstrated suboptimal performance in two-arm imitation learning, it was excluded from real test for simplicity.

Considering additional noise factors in the physical environment, such as joint and camera noise,  as well as the complexity of task design, visual input was obtained from three camera perspectives. 
Twenty successful trials were conducted to compute the success rate and average execution time.
The scoring criteria and requirements for three of the real-world tasks were shown in Fig. \ref{fig:7}.
\subsubsection{Results}
Table \ref{tab2} compared the performance of SRIL and ACT in real-world robotics experiments. Across three industry-like tasks, the SRIL demonstrated significant improvements in execution speed and success rates. In the Assembly task, it completed the task in 44.90\% of the time required by ACT, while in the Disassembly task, it achieved a 140\% speed increase and an 11.4\% higher success rate. For both tasks, its proactive approach and stable execution contributed to a 10.5\% average improvement in success rates.

In the conveyor belt dynamic gripping task, where ACT's mimicry strategy failed due to slow inference, while the proposed SRIL achieved a breakthrough with a 41.66\% success rate. This demonstrated its ability to improve execution speed along with improved accuracy, particularly in tasks requiring both precision and speed, highlighting its potential for real-world industrial applications.
\section{CONCLUSIONS}
This paper presented a novel strategy and training-inference framework designed to achieve rapid and efficient imitation learning in robotic systems. 
Motivated by human beings' subconscious behaviors, which did not continuously engage at every step, our method allows the robot’s imitation learning model to perform inference in a more subconscious manner. By reducing the frequency of model inference, we significantly accelerated robot performance while maintaining high accuracy. This strategy was further enhanced by preprocessing learning materials, enabling fine-grained and slow adjustments in detailed operations alongside the capability for rapid movement across long distance. Comprehensive experimental results in simulation and real robots demonstrated that the proposed framework not only increased the speed of task execution but also retained high success rates. This work offered valuable insights and a practical reference for advancing the deployment of imitation learning in real-world robotic applications. 

In future work, we will focus on enhancing the generalization and robustness of the proposed method. Specifically, we plan to optimize the inference strategy within the framework to dynamically adapt to varying task requirements, thereby further reducing inference frequency while maintaining precision in complex tasks. Additionally, we aim to explore more efficient data preprocessing and feature extraction techniques to improve the model’s ability to capture critical information in long-horizon tasks. Finally, we will evaluate the method’s performance in multi-task and dynamic environments to facilitate its broader deployment in real-world applications.

\bibliographystyle{IEEEtran}
\bibliography{reference}

\end{document}